\documentclass[a4j]{article}

\usepackage{jsaisig}
\usepackage{graphicx}

\usepackage[left=2cm, right=2cm, top=3cm, bottom=2.50cm]{geometry} 
\usepackage{helvet}
\usepackage{amsmath, amsfonts, amssymb} 
\usepackage[english]{babel}
\usepackage{graphicx,float}   
\usepackage{url}        
\usepackage{bm}        
\usepackage{multirow}
\usepackage{booktabs}
\usepackage[ruled,linesnumbered]{algorithm2e}  %algorithm code
\usepackage{indentfirst}
\begin{document}

\title{The Analysis about Building Cross-lingual Sememe Knowledge Base  Based on Deep Clustering Network}
\etitle{}

\author{Xiaoran Li
	\quad\quad\quad Toshiaki Takano
}

\affiliation{
		Adaptive System Lab, Shizuoka Institute of Science and Technology
		}

\abstract{
A sememe is defined as the minimum semantic unit of human languages. Sememe knowledge bases (KBs), which contain words annotated with sememes, have been successfully applied to many NLP tasks, and we believe that by learning the smallest unit of meaning, computers can more easily understand human language. However, Existing sememe KBs are built on only manual annotation, human annotations have personal understanding biases, and the meaning of vocabulary will be constantly updated and changed with the times, and artificial methods are not always practical. To address the issue, we propose an unsupervised method based on a deep clustering network (DCN) to build a sememe KB, and you can use any language to build a KB through this method. We first learn the distributed representation of multilingual words, use MUSE to align them in a single vector space, learn the multi-layer meaning of each word through the self-attention mechanism, and use a DNC to cluster sememe features. Finally, we completed the prediction using only the 10-dimensional sememe space in English. We found that the low-dimensional space can still retain the main feature of the sememes. 
}

\maketitle
\thispagestyle{empty}
%====================================================== Introduction
\section{Introduction}
\begin{figure}[h] 
\centering 
\includegraphics[width=0.4\textwidth]{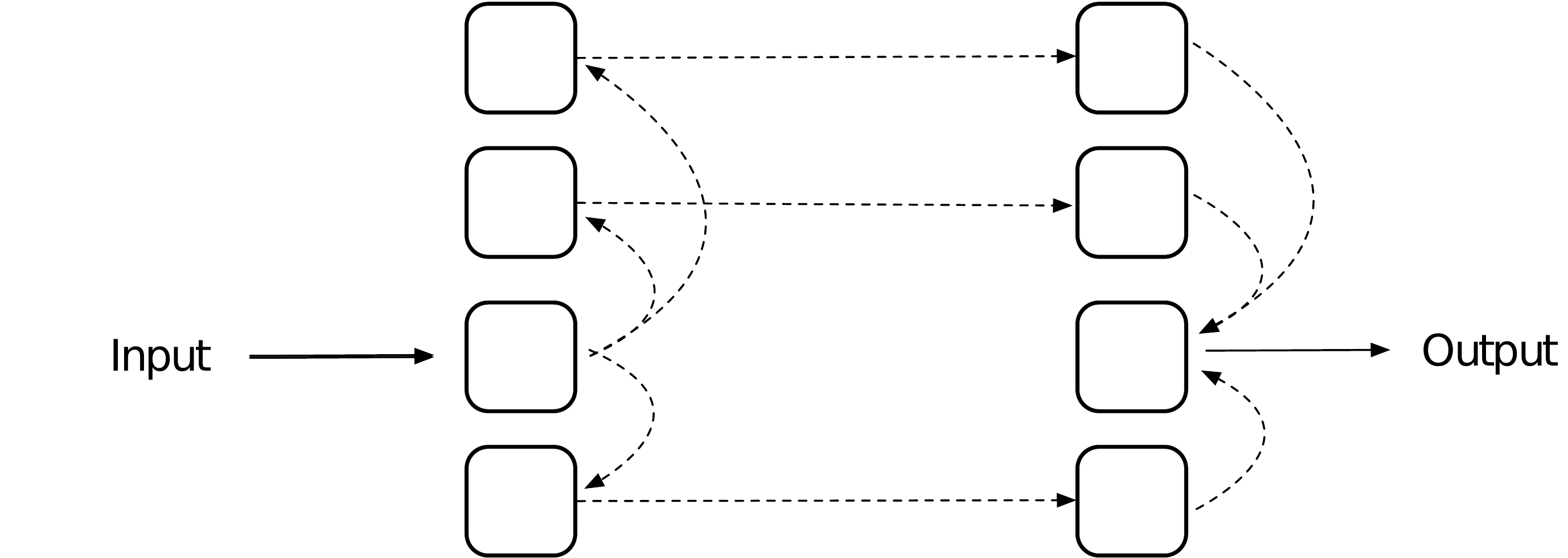} 
\caption{The input is the word embedding, and the output is the current meaning of the word. The dotted line represents the process of weighted summation of context words.} 
\label{Figure_1} 
\end{figure}

Because modern people use more diversified words, some words have more than one meaning. (e.g., the “apple” can be a kind of fruit or a company, and perhaps a banana company will appear in the future). However, No matter how a word changes its meaning, it is always composed of several primary and single meanings. In linguistics, a sememe (Bloomfield, 1926)\cite{Bloomfield_1926} is defined as the minimum semantic unit of human languages. Linguists believe that all languages have the same limited sememe space (Wierzbicka, 1996) \cite{Wierzbicka_1996}. Moreover, this construction method of sememes has been successfully used in various natural language processing (NLP) tasks. For example, the property that sememes can be combined into words has been applied in neural networks(Gu et al. 2018; Qi et al. 2019a)\cite{Gu2018}\cite{Qi2019a}. And there are also successful cases in sentiment analysis (Fu et al. 2013)\cite{Fu2013}, and semantic disambiguation (Duan, Zhao, and Xu 2007)\cite{Duan2007}.\\         
At this stage, people use manual annotation to build a sememe KB, such as HowNet(Dong and Dong, 2006)\cite{Dong_2006}, which uses about 2,000 language-independent sememes to annotate senses of over 100 thousand Chinese and English words. Alternatively, cooperate with the multilingual encyclopedic dictionary as BabelNet (Navigli and Ponzetto2012)\cite{Navigli_2012} to build a multilingual KB as (Qi et al. 2020)\cite{Qi2020}. However, semantics is an iterative system of continuous learning, HowNet is made purely by hand, and there is no interface for learning and evolution. And the meaning of words is a collection of multiple attributes. The construction of manual annotations will be biased by humans and ignore some word attributes.\\
To solve the flaws of manual labeling. In this paper, we tried to construct sememes in an unsupervised manner. Our method is motivated by 	(Arora et al. 2018)\cite{Sanjeev_2018} multiple senses of a word reside in linear superposition within the standard word embeddings (e.g., word2vec (Mikolov et al., 2013a)\cite{Mikolov_2013a}, and GloVe (Pennington et al., 2014)\cite{Pennington_2014}). Our idea is that if the surrounding words determine the meaning of a word, then the word's current meaning is determined by the weighted summation of the meanings of the surrounding words. According to this idea, we segment the meaning of each word through the self-attention mechanism based on word embedding (shown in Figure \ref{Figure_1}). We took the sub-meaning space of words by weighted summation of each word in each sentence as the original space of sememes for clustering. \\
However, the words after the weighted summation still have the dimensions of the original words, which is not suitable for clustering tasks. Because each word in each sentence in the corpus will generate a single word meaning, this will consume much memory for storage and calculation of clustering space.
Moreover, we guess that the sememe vector space with a single word meaning should have a relatively low dimensionality. However, although many data clustering methods have been proposed, conventional clustering methods usually have poor performance on high-dimensional data due to the inefficiency of similarity measures used in these methods. Furthermore, Word embedding has a highly complex data underlying structure. We want to find an effective method that can DR and cluster a large amount of high-dimensional data. \\
%-----------------------------------Figure_1
\\
%====================================================== Methodology
\section{Methodology}
In recent years, owing to the development of deep learning, deep neural networks (DNNs) can be used to transform the data into more clustering-friendly representations due to its inherent property of highly non-linear transformation (since DNNs can approximate any continuous mapping using a reasonable number of parameters (Hornik et al., 1989)\cite{Hornik_1989}). We hope to learn the low-dimensional minimum meaning of each word through the DNN to make it more friendly for clustering.  \\
\subsection{Deep Sememe Clustering}
\begin{figure}[h] 
\centering 
\includegraphics[width=0.5\textwidth]{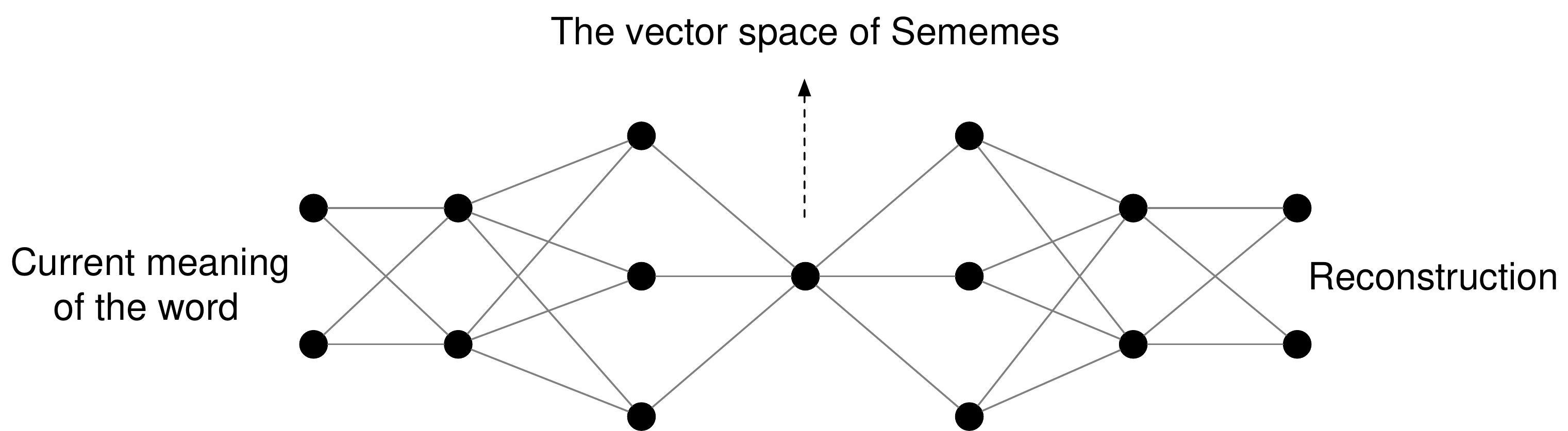} 
\caption{Construction of sememes based on DCN} 
\label{Figure_2} 
\end{figure}

In this article, we use the simplest and most effective deep clustering network (DCN) (Yang et al., 2017)\cite{Yang_2017}. DCN is one of the most remarkable methods in this field. It first learns the low-dimensional representation of the data by pre-training a DNN and then clusters the low-dimensional data as the initial value. Finally, the clustering effect is optimized through continuous iterative learning of low-dimensional space. It is fully compliant and can be applied to our sememe clustering(as shown in Figure \ref{Figure_2}). 
\subsubsection{Dimensionality Reduction}
DCN adopts an autoencoder (AE) (Vincent et al., 2010)\cite{Vincent_2010} to learn clustering-friendly representations. AE is a powerful method to train a mapping function based on DNNs, which ensures the minimum reconstruction error between the coder layer and data layer. Specifically, the approaches look into an optimization problem of the following form:
\begin{align} 
	L_{n} = \min _{\mathcal{W}, \mathcal{Z} } \sum_{i=1}^{N}\ell\left(\boldsymbol{g}\left(\boldsymbol{f}\left(\boldsymbol{x}_{i} ; \mathcal{W}\right) ; \mathcal{Z}\right), \boldsymbol{x}_{i} \right),
\end{align}
%-----------------------------------Figure_2
%-----------------------------------Figure_3
\begin{figure}[h] 
\centering 
\includegraphics[width=0.4\textwidth]{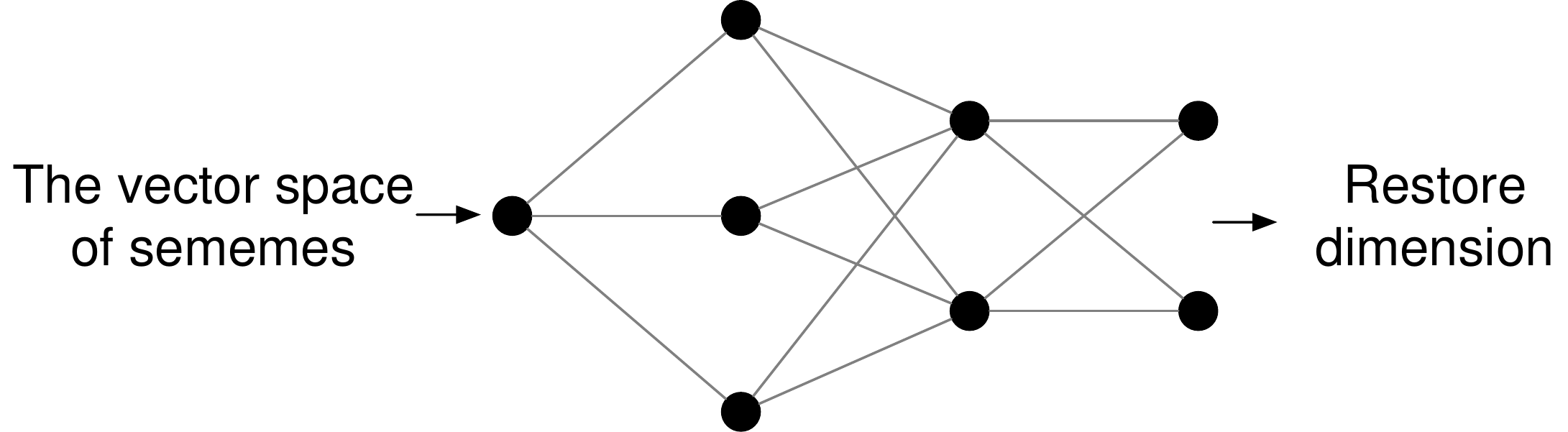} 
\caption{Restore data dimensions} 
\label{Figure_3} 
\end{figure}
where $\boldsymbol{f}\left( \cdot ; \mathcal{W}\right)$ denotes the nonlinear mapping function and $\mathcal{W}$ denote the set of parameters, i.e., 

$$
	\boldsymbol{f}(\cdot ; \mathcal{W}): \mathbb{R}^{M} \rightarrow \mathbb{R}^{R},
$$

$\boldsymbol{f}\left(\boldsymbol{x}_{i} ; \mathcal{W}\right)$ is the encoder network output given a set of data samples $\left\{\boldsymbol{x}_{i}\right\}_{i=1, \ldots, N}$, where $\boldsymbol{x}_{i} \in \mathbb{R}^{M}$ and $R \ll M$, Since the hidden layer usually has smaller dimensionality than the data layer, it can help find the most salient features of data. where  $\boldsymbol{g}\left( \cdot ; \mathcal{Z}\right): \mathbb{R}^{R} \rightarrow \mathbb{R}^{M}$ denotes the reconstruction function and $\mathcal{Z}$ denote the set of parameters. In the construction of sememes, it can help us map the low-dimensional sememes space to the original dimensional sememes space for evaluation (shown in Figure \ref{Figure_3}), we will explain in detail in the experimental part of this article. The function $\ell\left(\cdot\right): \mathbb{R}^{M} \rightarrow \mathbb{R} $ is a certain loss function that measures the reconstruction error. In the DCN, the least square loss is adopted as the reconstruction error, i.e., $\ell\left(\cdot\right)=\parallel\cdot\parallel^2_2$. \\

\subsubsection{Clustering}
	The optimization criterion of DNC is to connect DNN-based DR and clustering methods. The clustering method can be replaced, we only used the simplest and most effective K-means (Lloyd, et al., 1982)\cite{Lloyd_1982} in our research, which is also the clustering method adopted in the original work. The task of K-means is to group the $N$ data samples into $K$ categories by optimizing the following cost function:
\begin{align}
L_{c} = \min _{\boldsymbol{M}} & \sum_{i=1}^{N}\left\|\boldsymbol{f}\left(\boldsymbol{x}_{i} ; \mathcal{W}\right) -\boldsymbol{M} \boldsymbol{s}_{i}\right\|_{2}^{2}
\end{align}
	where $M  \in \mathbb{R}^{R \times K}$ is the sememe space we want to get, where $\boldsymbol{s}_{i} \in \mathbb{R}^{K}$ is the assignment vector of data point $i$ which has only one non-zero element, i.e., $\boldsymbol{M} \boldsymbol{s}_{i} \in \mathbb{R}^{R}$, $\boldsymbol{s}_{i}$ helps us to find the vector closest to the $\boldsymbol{f}\left(\boldsymbol{x}_{i} ; \mathcal{W}\right)$ in the $\boldsymbol{M}$ matrix, and then update the $\boldsymbol{M} \boldsymbol{s}_{i}$ to achieve our goal by continuously updating the $\boldsymbol{M}$ matrix. If we set $j$ as each element in $\boldsymbol{s}_{i}$, then the $\boldsymbol{s}_{i}$ vector can be defined as follows:
\begin{align}	
s_{i,j} \leftarrow\left\{\begin{array}{ll}
1, & \text { if } j=\mathop{argmin}\limits_{k=\{1, \ldots, K\}}\left\|\boldsymbol{f}\left(\boldsymbol{x}_{i} ; \mathcal{W}\right)-\boldsymbol{M}_{k}\right\|_{2} \\
0, & \text { otherwise. }
\end{array}\right.
\end{align}
For the above formula, we must first know the distribution of the $\boldsymbol{M}$ matrix to calculate, thus we must first obtain the low-dimensional representation of all the data through the AE, and obtain the initial matrix $\boldsymbol{M}$ by clustering all the low-dimensional data in advance distributed.

\subsubsection{Optimization}
	Finally, we use stochastic gradient descent (SGD) to iteratively optimize the encoder parameter $\mathcal{W}$, the decoder parameter $\mathcal{Z}$ and the sememe space $\boldsymbol{M}$ through the following formula:
\begin{align}
\min _{\mathcal{W}, \mathcal{Z}, \boldsymbol{M}} \sum_{i=1}^{N}\left(L_{n} + \lambda L_{c} \right)
\end{align}
	$\lambda \geq 0$ is a regularization parameter which balances the reconstruction error versus finding K-means friendly latent representations.	

\subsection{Cross-lingual Sememe}
\begin{figure}[h] 
\centering 
\includegraphics[width=0.5\textwidth]{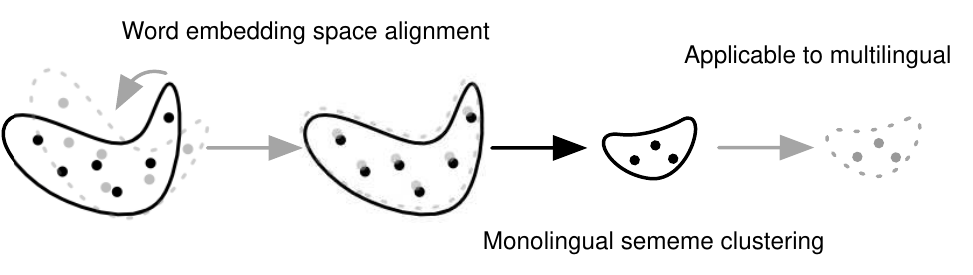} 
\caption{After aligning the word embedding space, only using monolingual word embedding for sememe clustering can be applied to other languages.} 
\label{Figure_4} 
\end{figure}

	Linguists have discovered that using a sememe space can be applied to any language (Wierzbicka, et al., 1996)\cite{Wierzbicka_1996}. thus we proposed whether we can learn the sememe space using only a common word embedding that has been aligned in a single vector space, we only need to learn a sememe space and it can be applied to all languages (as shown in Figure \ref{Figure_4}).
	In this way, we use the 30-word embeddings that have been aligned by MUSE (Conneau, et al., 2017) \cite{Conneau_2017} for learning cross-lingual word embeddings to enable model transfer between languages, e.g., between resource-rich and low-resource languages, by providing a common representation space. Its basic assumption is "words with similar semantics are close to each other in the word embedding space". With this, the two spaces can be brought closer by the following formula:
\begin{align}
W^{\star}=\operatorname{argmin}\|W X-Y\|_{\mathrm{F}}	
\end{align}
For the original space $X$ and the target space $Y$, it is hoped that a linear mapping $W$ can be found to make the converted $WX$ as close to $Y$ as possible. 
%-----------------------------------Figure_4

\section{Experiments}
\subsection{Pre-processing}
In the entire sememe clustering process, we must first obtain the original sememe space before clustering. Since we assume that each word has only one meaning in the current sentence, we need to find all the sentences containing this word in the entire corpus if we want to find all the meanings of a word. Therefore we need to create sentence indices for each word to find which sentence index contains this word in the corpus.
\subsubsection{Sentence Indices Dictionary (SID) }
\begin{figure}[ht] 
\centering 
\includegraphics[width=0.5\textwidth]{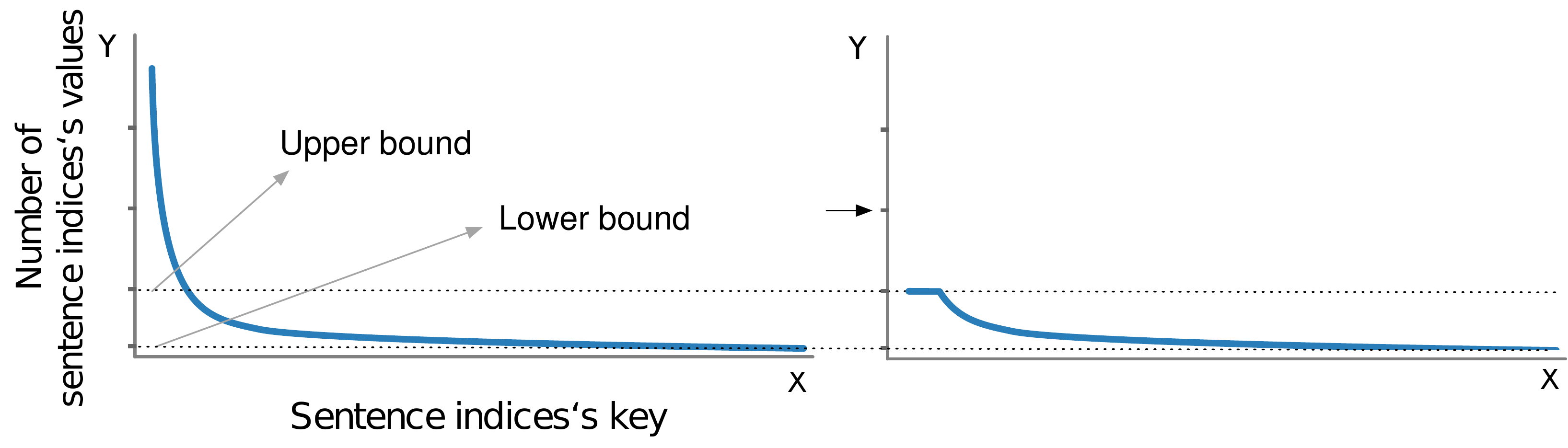} 
\caption{The illustration shows the distribution of the number of sentences. The X axis is the index of the word, and the Y axis is the number of sentences corresponding to the word. On the right is the distribution of the number of sentences after setting the upper and lower bounds.} 
\label{Figure_5} 
\end{figure}

We intercept 5GB in the Wikipedia corpus as a raw corpus for preprocessing and use SentencePiece (Kudo et al. 2018)\cite{Kudo_2018} 's BPE (Sennrich et al. 2016)\cite{Sennrich_2016} tokenizer to segment the raw corpus. We fixed the vocabulary of BPE to 200k. In terms of sentence segmentation, to make the meaning of each sentence as complete as possible, we did not directly use a fixed length for segmentation but used 42 kinds of symbols like“,.?!()”for segmentation. Moreover, because we use the self-attention mechanism to sum the proportions of all words in the sentence to get the current word meaning, if the sentence is too long, the weight of the important words will be assigned to the secondary words, which is difficult to get the obvious features of the current word meaning, thus we filter out sentences greater than 20 words and less than 2 words. Finally, we got 60 million qualified sentences to create sentence indices. \\
After constructing the sentence indices, we found that the number of sentences for specific words is enormous. Since our clustering network is to put all the meanings of all words into a space for clustering, If the sum of sentences for some words is close to half of the entire clustering space, the result of our clustering will tend to the meaning of these words. Therefore we have to balance the number of sentences in the sentence indices. To solve this problem, we implemented the following methods:\\
1. Delete stop words (We use stop words defined in NLTK for filtering.)\footnote{https://gist.github.com/sebleier/554280}\\
2. Set the upper and lower bounds for the number of sentences (shown in Figure \ref{Figure_5}). We set the upper and lower limits to 5k and two batch numbers respectively. If it exceeds 5k, randomly select 5k sentences from it, and if it is less than two batch numbers, discard the word.\\
%-----------------------------------Figure_5
3. We multiply the original data with an expansion coefficient $e \in \mathbb{R}$; and $e>0$ before the normalization function of self-attention, thus that it is not easy to lose important features when encountering long texts, In this experiment, we set $e=4$.\\
Finally, the number of sentences actually participating in clustering after being balanced is only 40 million. Note that the above parts are implemented locally, and will be used as input in DCN later.
%======================================================
\subsection{Clustering Process}
\subsubsection{Pre-training}
\begin{figure}[ht] 
\centering 
\includegraphics[width=0.4\textwidth]{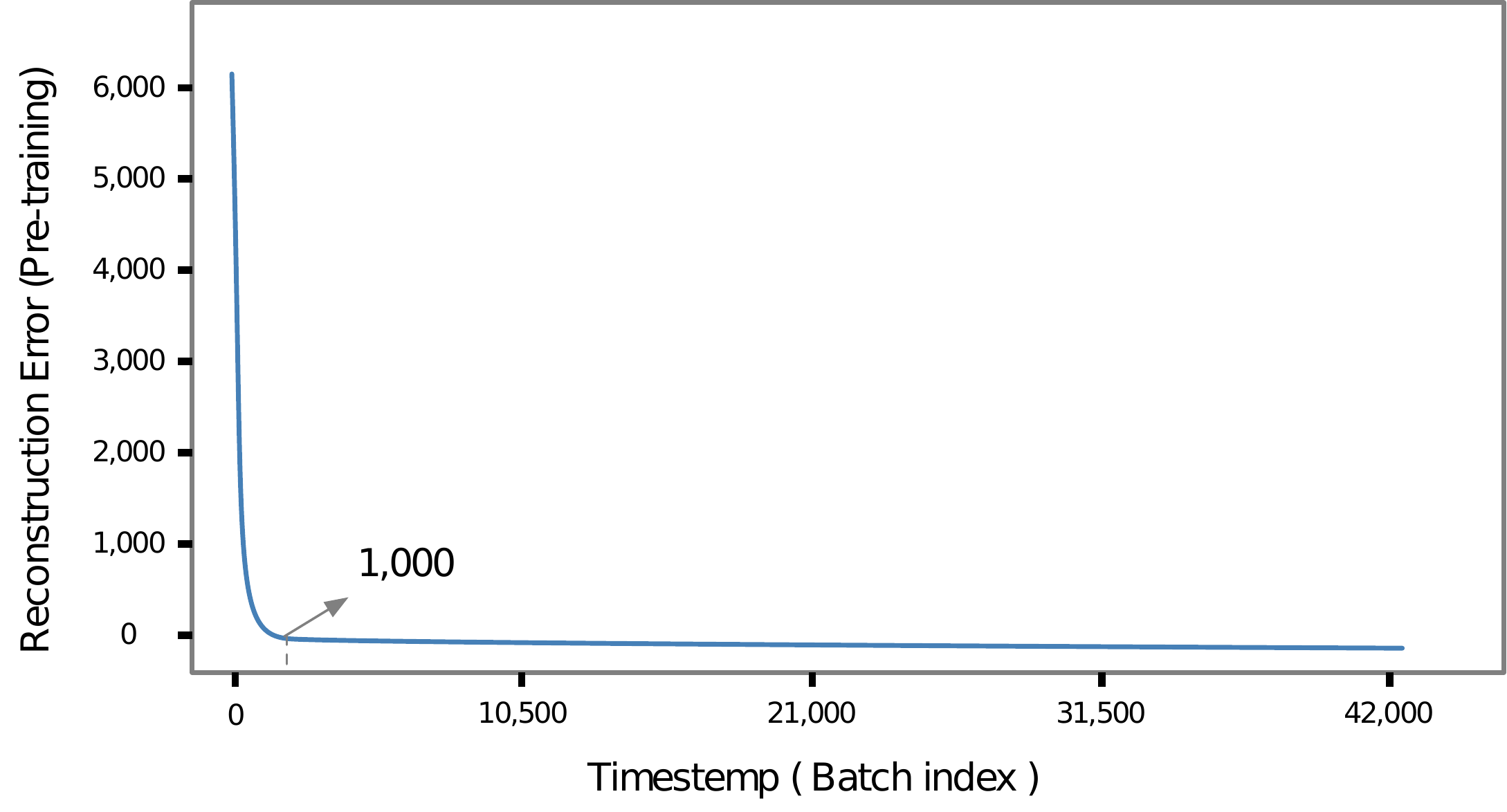} 
\caption{Reconstruction error of the autoencoder during the pre-training stage. (We set a batch size of 64 sentences and observed that it quickly converges and stabilizes within 1,000 batches)} 
\label{Figure_8} 
\end{figure}

The primary purpose of pre-training is to learn a low-dimensional cluster center matrix as the initialization of the sememe space $\boldsymbol{M}$ (shown in Algorithm \ref{algorithm_1}). Since the global distribution needs to be obtained during clustering, it is tremendous, but we cannot build this matrix in memory. Therefore, we adopted a sampling method when clustering. Each word only samples sentences of two batch numbers for clustering and loops ten times. Then the clustering results of ten times are clustered again to obtain the final cluster center initialization matrix. \\
\begin{algorithm}[h]
\caption{Sememe DCN Pre-training}\label{algorithm_1}
\KwData{Sentence indices dictionary $SID$, Word embeddings $WE$}
\KwResult{Initialization matrix $\boldsymbol{M}$}
\tcp{Train an autoencoder network}
\While{epoch}
	{
		\While{all vocabulary }{
			$batch  \leftarrow selfAttention(WE;SID)$\;
			\While{ $batch$ }{
				$\mathcal{W}^{\ast},\mathcal{Z}^{\ast} \leftarrow L_n(\mathcal{W};\mathcal{Z})$\;
			}
		}
	}
\tcp{Initialize the clustering space}
	$latent\_data \leftarrow$ new Array\;
	\While{all vocabulary }{
		$batch  \leftarrow selfAttention(WE;SID)$\;
		\While{ $batch$  }{
			$ latent\_data.append\left(\boldsymbol{f}\left(\boldsymbol{x}_{i} ; \mathcal{W}^{\ast}\right)\right)$\;
		}
	}
$\boldsymbol{M} \leftarrow kmeans\left(latent\_data\right)$\;
\end{algorithm}
In the experiment, we detection that the trained loss curve is jagged, Which may be caused by the uneven distribution of data samples. Since our method is to loop each word and perform the self-attention calculation on each word in turn. Although we scrambled the appearance order of the words before the loop, the loss was not smooth due to the inconsistency of word embeddings and sentences quality. Thus, according to the amount of memory used, we loop 500 words at a time and fully disrupt these words after doing self-attention (shown in Algorithm \ref{algorithm_1}). We defined the DCN-encoder size as [200, 200, 800, 10, 800, 200, 200], the input size is 300 dimensional aligned fasttext word embedding(Bojanowski et al.) \cite{Bojanowski_2017}. The learning rate is 0.003. And, the cluster centers are set to 2048, because we hope to make the cluster centers more dispersed while retaining the semantics of the sememe space to the greatest extent, we use t-SNE (Van der Maaten et al. 2008)\cite{van_2008} to reduce the dimensionality of the sememe space and analyze it by density, and found that 2k clusters can represent a rough meaning. If there are more clusters, the semantics will be more detailed. Figure \ref{Figure_8} presents the reconstruction error of the autoencoder during the pre-training stage.\\
%-----------------------------------Figure_8
%-----------------------------------Table_1
\begin{table*}[h]
	\centering
	\caption{Sememes cluster prediction in English}
	\scriptsize
	\begin{tabular}{ccccccc}
		\toprule 
		& 1st & 2nd &  3rd & 4th & 5th & 6th\\
		\midrule 
		projector & interfacing; & bluescreen; & wirelessly; & imageworks; &film; &windshields;\\
				&device;&screensavers; &handsets;&spotlighting;&movie;&plexiglass;\\
				&directionality&audio/visual &functionality&showcase&screenplay&windscreens\\
				&(0.03854)&(0.03646)&(0.03229)&(0.02812)&(0.02083)&(0.01667)\\
				\\
		woodland & grassy; & outcroppings; & porcupines; & lakeview; &vegetation; &northeast;\\
				&thickets;&bottomlands; &birds;&riverside;&habitats;&southwest;\\
				&vegetation&riverbeds &raccoons&park&habitat&northwesternmost\\
				&(0.07993)&(0.05168)&(0.04127)&(0.02564)&(0.02404)&(0.02143)\\
				\\
		stamen & prickles; &grasses;&laterally; &cottonwoods; & pinkish; & protruding;\\
				&fleshy;&berries;&tapering;&cattails; &yellowish;&shallowly;\\
				&leathery&shrubs&serrations&meadowsweet &brownish&lengthwise\\
				&(0.03854)&(0.03768)&(0.03401)&(0.02941)&(0.02574)&(0.01195)\\
				\\
		drone & pilotless; & spacecraft; & rabbuh; & melodically; &mobilizations; &gunnery;\\
				&refueling;&shuttlecraft; &zengi;&instrumentally;&stationing;&antiaircraft;\\
				&missile&spacelab &hashimi&sonically&forces&airlanding\\
				&(0.03559)&(0.03082)&(0.02908)&(0.02604)&(0.01823)&(0.01649)\\
				\\
		leaders &constitutionalists;&business;&political;&ideological;&election;&religious;\\
			     &politicize;&financial;&constitutionalist;&ideology;&elections;&religiousness;\\
			     &interventionist&investment&leftist&ideologies&party&trinitarianism\\
			     &(0.01723)&(0.01462)&(0.01262)&(0.01242)&(0.01182)&(0.01142)\\
			     \\
		marijuana&medication;&medication;&arrest;&constitutionality;&recourse;&prosecution;\\
			     &chronic;&medications;&suspects;&constitutionally;&overreaching;&conviction;\\
			     &hospitalization&diabetic&retribution&unconstitutionality&unconscionable&prosecute\\
			     &(0.08371)&(0.03051)&(0.02976)&(0.0253)&(0.02307)&(0.01823)\\
			      \\
		carrot&flavouring;&syrup;&grasses;&fruit;&buttered;&scared;\\
			     &sugared;&butter;&berries;&almonds;&sweetening;&scaredy;\\
			     &juice&juice&shrubs&blackcurrant&butter&crazy\\
			     &(0.07528)&(0.05256)&(0.04119)&(0.01989)&(0.01989)&(0.01847)\\
			      \\
		ozone&evaporation;&chemically;&transpiration;&gaseous;&chemically;&protrusions;\\
			     &evaporative;&peroxides;&particulates;&vaporization;&peroxides;&particulates;\\
			     &contaminants&nitrification&contaminants&hydrogen&solvents&concentrically\\
			     &(0.09809)&(0.04774)&(0.0434)&(0.03559)&(0.02865)&(0.02778)\\

		\bottomrule
	\end{tabular}%
	\label{table_1}%
\end{table*}%
\subsubsection{Fine-tuning And Results}
In the Fine-tuning stage, we use the original DCN to optimize the sememe space, and we found that the reconstruction loss of the autoencoder in the Fine-tuning phase is consistent with the pre-training phase and there is no downward trend. 
Finally, we predicted the sememes of some words and their probabilities and used the existing English $SID$ to search for sememes in other languages (shown in Table \ref{table_1}). Through the table, we found that sememe can express certain characteristics of words. The decimals in the table represent the probability of this sememe in the overall sememe space (Only the 6 with the highest probability are listed). The sememe we calculated is just a vector of the original word embedding space, it is not a word, therefore we find the nearest three words by cosine similarity to represent the meaning of sememe. Since the sememes predictions in other languages depend entirely on the quality of English sememes and the degree of alignment with English, and you can use the data set that MUSE has already trained, thus we will not show the relevant results of other language predictions here. \\
%======================================================
\section{Conclusion}
In this paper, we simply use DCN to analyze the effect of sememe clustering. We compress the size of sememe to 10 dimensions. Although most of the information is lost, the main meaning can still be retained. Although in Table \ref{table_1} we only show some good prediction results, it proves that our prediction is correct. we believe this is a good beginning.\\
In the future, we will try other methods to improve prediction performance, e.g., we can use the pivot language method (by intervening in the language of other language families for translation and then performing the reverse translation.) We hope that the Cross-language family translation can retain the basic meaning of semantics to reduce the sememe noise of a single language. Moreover, because we hope that our sememes research can contribute more to natural language understanding (NLU) that can truly allow computers to parse semantic information, we will combine NLU tasks and improve their performance by using sememes in future research.

\end{document}